# Generalizability of Machine Learning Models: Quantitative Evaluation of Three Methodological Pitfalls

Farhad Maleki[1,‡], Katie Ovens[1,‡], Rajiv Gupta[3], Caroline Reinhold[2,4], Alan Spatz[5], and Reza Forghani[2,4,6,*]

[1]Department of Computer Science, University of Calgary, AB, Canada
[2]Augmented Intelligence & Precision Health Laboratory (AIPHL), Department of Radiology and the Research Institute of the McGill University Health Centre, McGill University, Montreal, QC, Canada
[3]Department of Radiology, Massachusetts General Hospital, Boston, MA, USA
[4] Montreal Imaging Experts, Montreal, QC, Canada
[5] Division of Pathology, Jewish General Hospital, Montreal, QC, Canada
[6] Radiomics and Augmented Intelligence Laboratory (RAIL), Department of Radiology and the Norman Fixel Institute for Neurological Diseases, University of Florida College of Medicine, Gainesville, FL

[‡]Authors contributed equally.
[*]Corresponding author: r.forghani@ufl.edu

## Abstract

**Purpose:** Despite the potential of machine learning models, the lack of generalizability has hindered their widespread adoption in clinical practice. We investigate three methodological pitfalls: (1) violation of independence assumption, (2) model evaluation with an inappropriate performance indicator or baseline for comparison, and (3) batch effect. **Materials and Methods**: Using several retrospective datasets, we implement machine learning models with and without the pitfalls to quantitatively illustrate these pitfalls' effect on model generalizability. **Results:** Violation of independence assumption, more specifically, applying oversampling, feature selection, and data augmentation before splitting data into train, validation, and test sets, respectively, led to misleading and superficial gains in F1 scores of 71.2% in predicting local recurrence and 5.0% in predicting 3-year overall survival in head and neck cancer as well as 46.0% in distinguishing histopathological patterns in lung cancer. Further, randomly distributing data points for a subject across training, validation, and test sets led to a 21.8% superficial increase in F1 score. Also, we showed the importance of the choice of performance measures and baseline for comparison. In the presence of batch effect, a model built for pneumonia detection led to F1 score of 98.7%. However, when the same model was applied to a new dataset of normal patients, it only correctly classified 3.86% of the samples. **Conclusions**: These methodological pitfalls cannot be captured using internal model evaluation, and the inaccurate predictions made by such models may lead to wrong conclusions and interpretations. Therefore, understanding and avoiding these pitfalls is necessary for developing generalizable models.

## Introduction

Medical images are widely used for diagnosis and treatment planning. Manual qualitative evaluation of these images by domain experts is the most common method for analyzing such data. This process is time-consuming and prone to inter-observer and intra-observer variabilities (1, 2). Human interpretation also may not fully leverage many quantitative features that may not be apparent to the naked eye. Machine learning (ML) and deep learning (DL) have great potential for supplementing and augmenting expert human assessment by acting as a clinical assistant or decision support tool (3-9).

Despite a large body of published work on applications of ML and DL in medicine, very few are clinically deployed (10). Lack of generalizability of trained models is an important reason hindering the widespread deployment of these methods in clinical settings (10). Factors that affect generalizability include technical variations and lack of standardization in medical practice, differences in patient demographics from one center to another, patient genotypic and phenotypic characteristics, and differences in tools and methodologies used for medical data processing and model development (11).

Multiple guidelines for conducting and presenting research to ensure rigor, quality, and reproducibility of ML and DL models have been published (12-16). QUADAS and its extension QUADAS-2 were developed by Whiting *et al.* for a systematic review of diagnostic studies (15). QUADAS-2 assesses the risk of bias in patient selection, index test, reference standard, and flow and timing of a diagnostic study to ensure generalizability. Wolff *et al.* designed PROBAST as a series of questions to facilitate systematic review and assessment of potential bias in clinical prediction models (16). Collins *et al.* developed the TRIPOD guideline to encourage transparency in reporting prediction models (12). TRIPOD contains recommendations for expected content and characteristics of abstract, introduction, methods, results, and discussion sections of scientific papers on ML and DL. Mongan *et al.* published the CLAIM checklist to aid authors and reviewers with best practices in artificial intelligence research in medical imaging (14). Similar to TRIPOD, CLAIM provides high-level recommendations for preparing scientific manuscripts but is focused on medical imaging. Among these works, CLAIM is the most widely used checklist for artificial intelligence (AI) in medical imaging.

The aforementioned guidelines mainly focus on reporting and reproducibility aspects of research findings. However, they offer minimal to no guidance regarding good methodological practices in medical applications of machine learning. Even when suggested guidelines are followed and the results are reproducible, there is still a risk of methodological errors in the study design and execution. Only a limited number of explicit technical guidelines exist for avoiding methodological mistakes that lead to a lack of generalizability of ML and DL applications (10). In addition, they are often presented in a manner that is not readily accessible to practitioners in the medical domain. Clear, scientifically-backed guidelines are essential to promote the development of generalizable ML and DL models that may be clinically deployed.

In this paper, we identify and experimentally investigate the following three major categories of methodological errors in developing ML and DL models: (1) violation of independence assumption, (2) the use of inappropriate performance indicators for model evaluation, and (3) the introduction of batch effect. We also provide guidelines for avoiding these pitfalls.

## Materials and Methods

### Datasets

To show that the aforementioned methodological pitfalls are not specific to one data modality, in this study, we use several imaging modalities.

*Head and Neck Squamous Cell Carcinoma (HNSCC) CT dataset*

We used a CT dataset of 137 head and neck squamous cell carcinoma (HNSCC) patients treated by radiotherapy (17, 18). Hereafter, we refer to this dataset, which consists of pre-treatment CT scans, as "HNSCC". The dataset is available from The Cancer Imaging Archive (TCIA) (17-19). Table 1 provides a summary of the clinical endpoints of the HNSCC dataset.

*Table 1. Distribution of clinical features for the HNSCC dataset.*

|  | Larynx (49 patients) | Oropharynx (88 patients) |
|---|---|---|
| Sex (Female: Male) | 5:44 | 21:67 |
| Age | 65.8 ± 8.8 | 59.8 ± 7.9 |
| T Stage (1: 2: 3: 4) | 21:4:11:13 | 14:28:13:33 |
| N Stage (0: 1: 2: 3) | 37:2:9:1 | 23:14:49 :2 |
| 3-Year Overall Survival (No: Yes) | 11:38 | 27:61 |
| Local Recurrence (No: Yes) | 37:12 | 76:12 |
| Distant Metastasis (No: Yes) | 48:1 | 81:7 |

*Lung CT Dataset*

We used the images from the Lung CT Segmentation Challenge 2017, extracted from TCIA, containing 120 CT scan series from 60 patients (19-21). We use this dataset to demonstrate methodological pitfalls related to performance metrics for segmentation.

*Digital Histopathology Dataset*

We used a pathology dataset containing 143 hematoxylin and eosin (H&E)-stained formalin-fixed paraffin-embedded whole-slide images of lung adenocarcinoma provided by the Department of Pathology and Laboratory Medicine at Dartmouth-Hitchcock Medical Center (22). The dataset contains five histopathological patterns: solid (51 slides), lepidic (19 slides), acinar (59 slides), micropapillary (9 slides), and papillary (5 slides). We used the 110 slides from patients with solid and acinar predominant histopathological patterns as they are numerous and relatively balanced.

Due to the high resolution of the histopathology images, it is computationally impractical to analyze them as a whole image (23). Therefore, we first downscaled each image by a factor of 4. Then, using color thresholding, we extracted the foreground, i.e., the tissue segments on each slide. Next, for each image, we extracted random patches sized 1024 by 1024 pixels. Patches with 75% or more background were excluded during the patch extraction process. The patch extraction process continued until 200 patches were extracted from each image resulting in 22000 patches. Figure 1 illustrates an example whole-slide image as well as a selection of random patches.

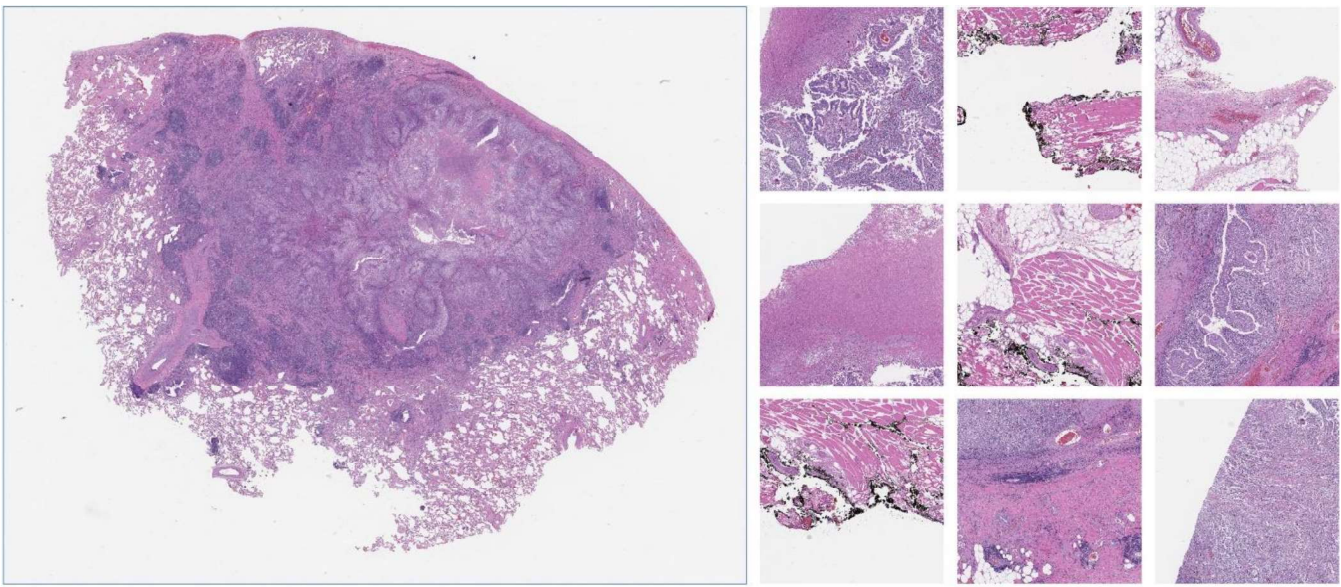

*Figure 1. Example of a whole-slide image from the Dartmouth-Hitchcock Medical Center lung adenocarcinoma dataset with nine example patches extracted from the image.*

*Chest X-ray Datasets*

To demonstrate the impact of batch effects, we used two X-ray datasets: 8851 normal chest X-rays with no findings from the Radiological Society of North America (RSNA) pneumonia detection challenge dataset, which is available on Kaggle, and a chest X-rays dataset from Kermany *et al. (24)*, which included 1349 normal X-rays with no findings and 3883 X-rays demonstrating pneumonia in pediatric patients. Figure *2* illustrates samples from each dataset.

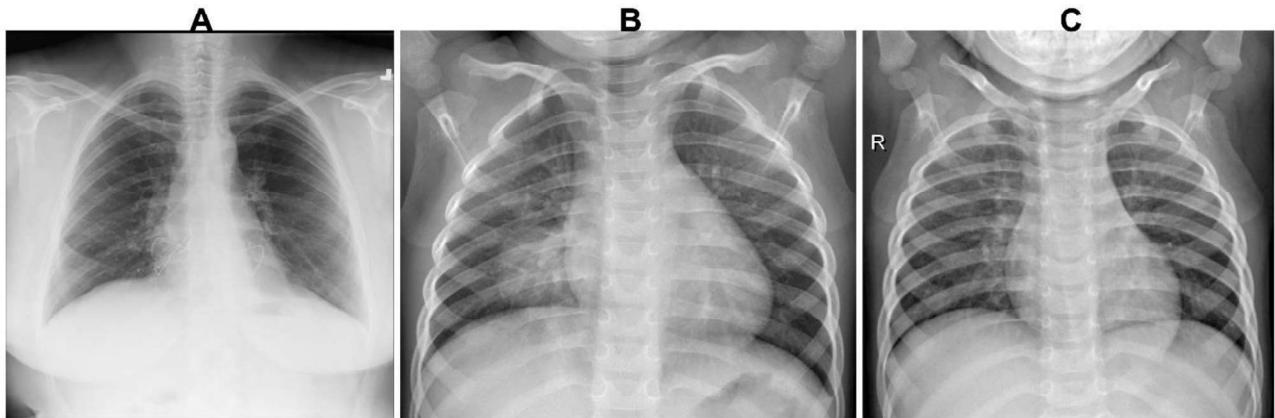

*Figure 2. Three example images from the chest X-ray dataset. A) Chest X-ray image of a normal (no findings) adult subject from the RSNA pneumonia detection challenge dataset, B) Chest X-ray image of pediatric pneumonia, and C) Chest X-ray image of a normal pediatric (no findings) subject.*

## Experiments

This section presents the experiments designed to investigate the effect of the three major categories of methodological errors on model generalizability.

### *(1) Breaking the Assumption of Independence*

ML and DL approaches assume that data used for model training and evaluation are independent and identically distributed. Based on this assumption, samples used for model training and evaluation have the same probability distribution and are mutually independent. To develop machine learning models, it is a common practice that the available data is split into training, validation, and test sets. The training set is used to learn model parameters, the validation set is used to select model hyper-parameters, and the test set is used to provide an unbiased estimate of model generalization error. However, the validity of this design is contingent on the assumption of independence. To provide an unbiased estimate of the generalization error, the test set needs to be independent of the training and validation sets.

Adhering to the assumption of independence is essential for developing generalizable models. Data used for model training should be independent of the data used for model evaluation. This assumption can be violated when some data that are not expected to be available in the prediction/evaluation phase is used for model training. This phenomenon is referred to as data leakage. We investigate the impact of different designs and executions of four common practices that could lead to data leakage and violation of the assumption of independence: (I) over-sampling, (II) data augmentation, (III) using several data points from a patient, and (IV) feature selection.

*(I) Over-sampling:* Class imbalance happens when there is a substantial difference between the number of samples from one class versus the others. Often, models developed using imbalanced datasets tend to undermine minority classes and focus on the majority classes. Oversampling is a technique used to alleviate this challenge by extracting a larger number of samples using sampling with replacement from the original minority classes to artificially increase the number of samples from the minority class(es). Class imbalance is common in medical imaging datasets due to factors such as the rarity of some diseases and difficulties in imaging certain conditions (25).

To quantitatively show how applying oversampling before splitting data into training, validation, and test sets could affect model generalizability, we evaluate two binary classifiers—as described in Appendix A—for predicting local recurrence in head and neck cancer by employing a radiomics approach using the HNSCC dataset. The first model (model A) is developed by conducting oversampling before data splitting. The second model (model B) is developed by conducting oversampling after data splitting. The only difference between these models is the order of applying the oversampling step.

To achieve statistically reliable results, we repeat this process 100 times and calculate the F1 score as the performance measure. We then use the Wilcoxon Rank-Sum test to assess if there is a statistically significant difference between the performance measures derived from model A versus model B.

*(II) Data augmentation:* Data augmentation refers to computational methods used to generate new data points from the existing ones. Data augmentation is commonly used when developing ML and DL models for image analysis, and it has been shown to improve the performance and generalizability of the resulting models. The use of data augmentation is essential in medical image analysis, where developing large-scale datasets is often impractical (26).

To quantitatively show how applying data augmentation before splitting data into training, validation, and test sets could affect model generalizability, we develop two DL-based binary classifiers—as described in Appendix B—for distinguishing solid and acinar predominant histopathological patterns in patients with lung adenocarcinoma. The first model (model C) is developed by conducting data augmentation before data splitting. The second model (model D) is developed by conducting data augmentation after data splitting. For these models, every component of the model building and evaluation pipeline is kept the same, other than the order of applying the data augmentation step.

*(III) Using several data points from a patient:* We experimentally investigate how distributing data points for a patient across training, validation, and test sets could impact model generalizability. Using a pathology dataset, we build two DL-based binary classifiers—as described in Appendix B—for distinguishing solid and acinar predominant histopathological patterns for patients with lung adenocarcinoma. Model E is developed by randomly distributing the image patches across training, validation, and test sets. Model F is developed by assigning image patches for each patient to either training, validation, or test sets. Everything else is kept the same for these models.

*(IV) Feature selection:* To demonstrate the impact of applying feature selection before splitting data into training, validation, and test sets on model generalizability, using the HNSCC dataset, we develop two binary classifiers—as described in Appendix A—to predict overall survival using a radiomics approach. The first model (model G) is developed by conducting feature selection before data splitting. The second model (model H) is developed by conducting feature selection after data splitting. For the sake of accurate comparison, all other steps are kept the same for these models.

To achieve statistically reliable results, we repeat this process 100 times and calculate the F1 score as the performance measure. We then use the Wilcoxon Rank-Sum test to assess if there is a statistically significant difference between the performance measures derived from model G versus model H.

*(2) Evaluating Models with an Inappropriate Performance Indicator or Baseline for Comparison*

The choice of quantitative measures to reflect the utility of a model is essential for developing predictive models. In addition, setting the baseline for acceptable performance is required to determine if a model could be useful and deployable in real-world settings.

Accuracy, which represents the proportion of samples correctly classified by a model, is commonly used to measure the performance of classification models. Using a model for distant metastasis prediction in the HNSCC dataset, we show that using accuracy as a performance indicator for imbalanced datasets—which are common in the medical domain—could lead to an erroneous interpretation of model performance.

We empirically investigate how an inappropriate choice of baseline for performance comparison could result in misleading outcomes and erroneous conclusions. The performance of segmentation models is commonly measured using IoU (Intersection over Union) and Dice scores. Dice score is a measure of relative overlap and is defined as follows:

$$Dice(X, Y) = \frac{2 \, \|X \cap Y\|}{\|X\| + \|Y\|}$$

where $X$ and $Y$ are two segmentations, e.g., the ground truth and the model prediction. Also, IoU is calculated as follows:

$$IoU(X, Y) = \frac{\|X \cap Y\|}{\|X \cup Y\|}$$

Dice score and IoU result in values between 0 and 1, where a value of 1 represents a perfect overlap between $X$ and $Y$, and 0 represents no overlap.

To show the role of setting a baseline expectation in evaluating the result of a segmentation model, we develop a threshold-based approach to segment air inside the body for samples in the lung CT dataset as a proxy for lung segmentation to serve as a baseline model. Any sophisticated model for lung segmentation should outperform such a baseline.

We use any voxel with a Hounsfield unit less than -400 as air. Then we remove the segment of air outside the bodies of the patients. This results in segmenting air within the body, which is used as an erroneous proxy for lung segmentation. We compare the performance of this simple model—which could be used as a simple baseline—with the ground truth (lung contours). Any model with a performance less than such a baseline model should be considered irrelevant.

*(3) Batch Effects*

A batch effect happens when data that come from several sources are aggregated to develop a larger dataset, and the class distribution samples from these sources substantially vary. For example, suppose all malignant tumors in a cohort were scanned on MRI scanner X and all benign tumors on MRI scanner Y. In that case, the model may have learned to differentiate the malignant from the benign tumors not based on intrinsic tumor characteristics but rather due to MRI scanner-attributed differences. We hypothesize that batch effect can be used by ML and DL models to superficially boost performance measures, and the resulting model could learn factors that characterize each batch rather than the condition under study. In order to experimentally investigate how batch effects could impact model generalizability, we simulate a dataset with batch effect by extracting pneumonia samples from the dataset by Kermany *et al.* (24) and the normal samples (i.e., images with no findings) from the RSNA dataset. Since the pneumonia samples come from a pediatric population and the normal samples come from an adult population, there is a batch effect present in this dataset characterized by differences in anatomical structures and body position. Hereafter we refer to this dataset as Batch X-Ray. We train a model (model I) on the Batch X-Ray dataset. Then we test this model on an external dataset of normal chest X-Ray images from Kermany *et al.* (24) to show how the batch effect could affect the model generalizability.

# Results
## (1) Violation of independence assumption

Table 2 shows the effect of incorporating the described methodological pitfalls versus when these pitfalls are avoided. For model A, oversampling was conducted before splitting data. For model B, oversampling was applied after splitting data. These models were run 100 times, and the average performance measures were reported. The results showed a substantial and statistically significant gap between the performance of these two models. While model B—the correct approach—showed poor performance, model A seemed to offer promising results. The Wilcoxon Rank Sum tests indicated that incorrect application of oversampling leads to a statistically significant but superficial increase in model performance (statistic=-12.22, p-value <0.001) for predicting local recurrence in head and neck cancer.

Models C, D, E, and F were built for distinguishing solid and acinar predominant histopathological patterns in patients with lung adenocarcinoma. Since developing DL models is computationally expensive, and due to the relatively larger dataset sizes for model development, we followed the common practice of reporting the results for one trained model, unlike the radiomics models A, B, G, and H, where we repeated model building for 100 times and reported average performance measures. In model C, data augmentation was applied before splitting data and in model D after splitting data. The results showed a superficial boost in the performance measures for model C, while model D showed poor performance.

The effect of breaking the independence assumption by distributing data points for a patient across training and test sets is demonstrated with models E and F. For model E, data points have been randomly distributed across training, validation, and test sets. Therefore, data points for a patient could appear in both training and test sets. For model F, the independence assumption has been preserved by assigning data points for each patient to either training, validation, or test set. Note that model F is the same as model D—i.e., the correct approach for distinguishing solid and acinar predominant histopathological patterns for patients with lung adenocarcinoma. These results indicate that distributing data points of patients across training and test sets leads to a superficial boost in the performance of model E. In contrast, the performance measures for model F are substantially lower than that of model E.

Table 2 also shows how applying feature selection prior to splitting data into training, validation, and test sets could lead to violation of the independence assumption. For model G, feature selection has been conducted before splitting data and for model H after splitting data. The result indicates that applying feature selection before splitting data into training, validation, and test sets could lead to a superficial boost in the model performance for predicting 3-year overall survival in head and neck cancer. The Wilcoxon Rank Sum tests indicated that this misleading boost is statistically significant (statistic=-5.87, p-value<0.001).

*Table 2. Performance measures of models built when incorporating methodological pitfalls versus performance measures of models built when the pitfalls are avoided. Model A was built with oversampling performed before splitting data into training, validation, and test sets and model B was built with oversampling performed after splitting data. Both models were trained to predict local recurrence in head and neck cancer using the HNSCC dataset. For model C, augmentation was performed before splitting data into training, validation, and test sets and for model D after splitting data. Model E was developed by randomly distributing all data points across training, validation, and test sets. For model F, all data points for a patient were assigned to either training, validation, or test set. Models C, D, E, and F were trained to distinguish solid and acinar predominant histopathological patterns in patients with lung adenocarcinoma. Since developing DL models is computationally expensive, and due to the relatively larger dataset sizes for model development, for these models, we followed the common practice of reporting the results for one trained model. In model G, feature selection was conducted before splitting data into training, validation, and test sets and in model H after splitting data. Both models were built to predict 3-year overall survival using the HNSCC dataset. Note that for models A, B, G, and H, performance measures were acquired as the average of performance measures for 100 runs of the model development pipeline. For these models, we also report the standard deviation of the performance measures. The Wilcoxon Rank Sum tests indicated that incorrect application of oversampling and feature selection leads to statistically significant superficial increases in model performance for predicting local recurrence (statistic=-12.22, p-value <0.001) and 3-year overall survival (statistic= -5.87, p-value<0.001) in head and neck cancer.*

| Model | Algorithm | Approach | Accuracy | Precision | Recall | F1 |
|---|---|---|---|---|---|---|
| A | Random forest | Incorrect oversampling | 0.85±0.05 | 0.82±0.07 | 0.89±0.07 | 0.85±0.05 |
| B | Random forest | Correct oversampling | 0.72±0.06 | 0.16±0.17 | 0.15±0.14 | 0.14±0.11 |
| C | ResNet-50 | Incorrect data augmentation | 0.999 | 0.999 | 0.999 | 0.999 |
| D | ResNet-50 | Correct data augmentation | 0.659 | 0.530 | 0.548 | 0.539 |
| E | ResNet-50 | Incorrect distributing data points of patients | 0.788 | 0.799 | 0.719 | 0.757 |
| F | ResNet-50 | Correct distributing data points of patients | 0.659 | 0.530 | 0.548 | 0.539 |
| G | Random forest | Incorrect feature selection | 0.72±0.07 | 0.80±0.06 | 0.81±0.08 | 0.80±0.05 |
| H | Random forest | Correct feature selection | 0.64±0.07 | 0.76±0.06 | 0.76±0.06 | 0.75±0.05 |

## (2) Model evaluation with an inappropriate performance indicator or baseline for comparison

Assume a naive model that predicts all samples as non-distant metastasis. This model achieves an accuracy of 94% ($\frac{48 + 81}{48 + 1 + 81 + 7}$) for our HNSCC dataset. However, such a model has no medical utility as it achieves a recall of zero and fails to diagnose any distant metastasis. This example shows that using accuracy for highly imbalanced datasets might lead to developing models that cannot be deployed in a clinical setting.

Figure 3 illustrates the ground truth segmentation as well as predicted segmentation of a randomly chosen image from the lung CT dataset, where the prediction has been made by a simple baseline model that detects air inside the body. While the predicted segmentation for lung is not medically acceptable, it achieves a Dice score of 0.94 and an IoU of 0.88. Table 3 shows the summary of the Dice score and IoU for the result of a simple model detecting air within the body on the lung CT dataset as a baseline for lung segmentation. The results of this simple model achieved a high Dice score and IoU, while a visual inspection reveals the inferiority of the segmentation from a medical perspective. Therefore, models with performance measures lower than this baseline model should not be utilized even if their measured performance seems high.

Table 3. Summary of Dice and IoU score for a simple model detecting air inside the body as an estimate for lung segmentation in chest CT scans of the Lung Segmentation Challenge 2017. The summary highlights the mean, minimum, and maximum Dice score and IoU between the prediction made by the thresholding approach and the ground truth resulting from manual segmentation. Considering the thresholding approach as a computationally efficient but not medically reliable baseline for lung segmentation, any sound lung segmentation model should at least outperform this baseline. For example, this means that a deep learning model with a Dice score of 0.92 should be considered a low-performing model.

|      | Mean | Min  | Max  | Standard Deviation |
|------|------|------|------|--------------------|
| Dice | 0.92 | 0.84 | 0.97 | 0.02               |
| IoU  | 0.86 | 0.72 | 0.94 | 0.04               |

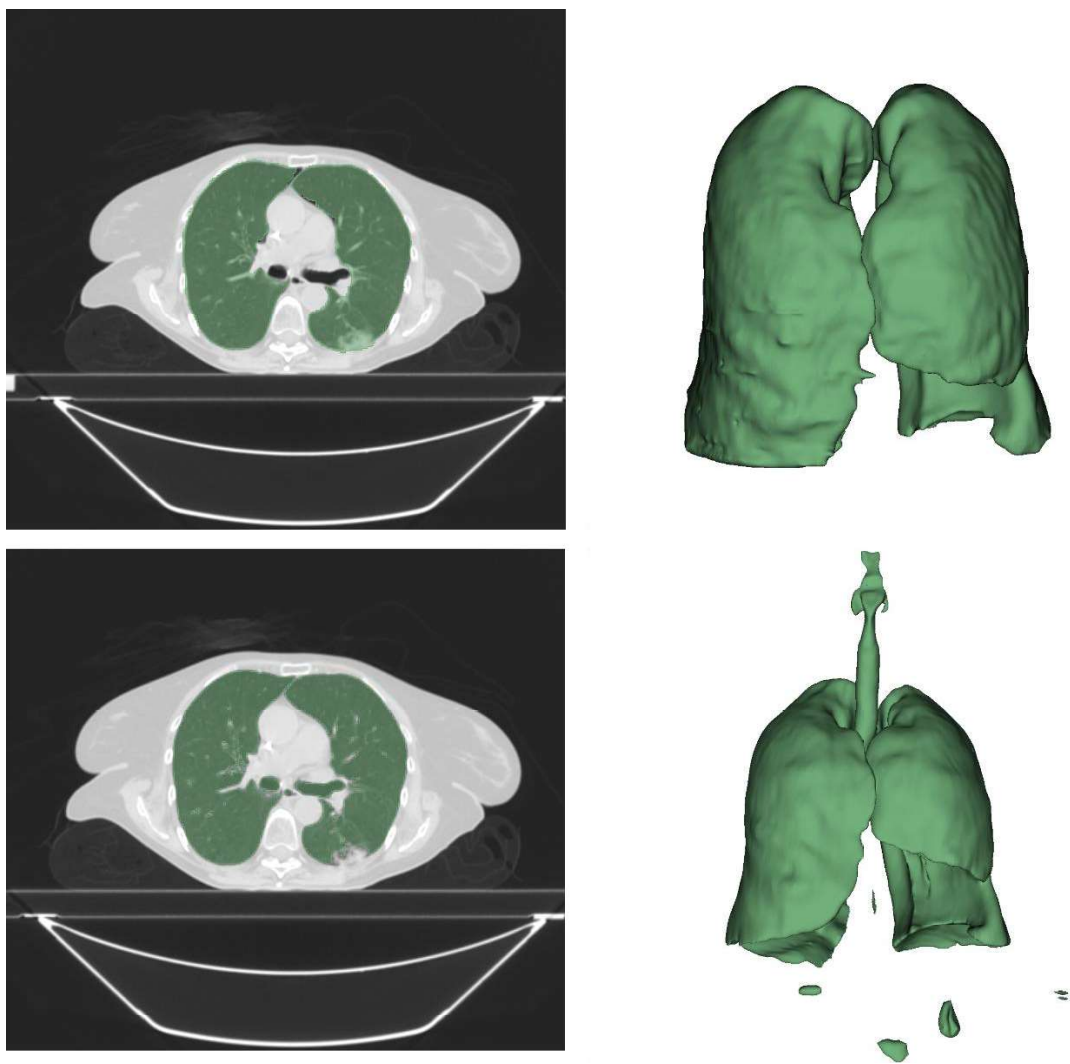

Figure 3. CT images of the lung with segmentation (shown in green). The top-left image shows the ground truth mask overlaid on a slice of a chest CT manually contoured by a radiologist. The top-right image illustrates the 3D volume of the manual contour. The bottom-left image illustrates a slice of the predicted segmentation mask overlaid on its corresponding slice of the CT image. The prediction has been made by a simple baseline model that detects air within the body. The bottom-right image shows the 3D volume of the prediction made by the baseline model. The segmentation includes air in the body (including trachea and bowel gas) using the provided chest CT scan. This highlights that for large volumes, such as lung, a high Dice score might not be indicative of high-quality segmentation.

## (3) Batch effect

Our results also showed that batch effects could be a substantial barrier to model generalizability. We observed that model I—which was trained and tested on a dataset with batch effect (Batch X-Ray dataset)—achieved accuracy, precision, recall, and F1 score of 0.997, 0.979, 0.995, and 0.987, respectively. However, when this model was applied to the normal pediatric chest X-ray samples from the dataset by Kermany *et al*. (24), only 3.86% of samples were classified correctly as normal.

The attribution of each pixel of an image on the model prediction for that image can be calculated using Integrated Gradient method (27). Figure 4 overlays the attribution values for each pixel of an X-Ray of a normal pediatric sample. As depicted in Figure 4, the pneumonia prediction model trained using the Batch X-Ray dataset focuses on anatomical structures and body position rather than image characteristics in the lung.

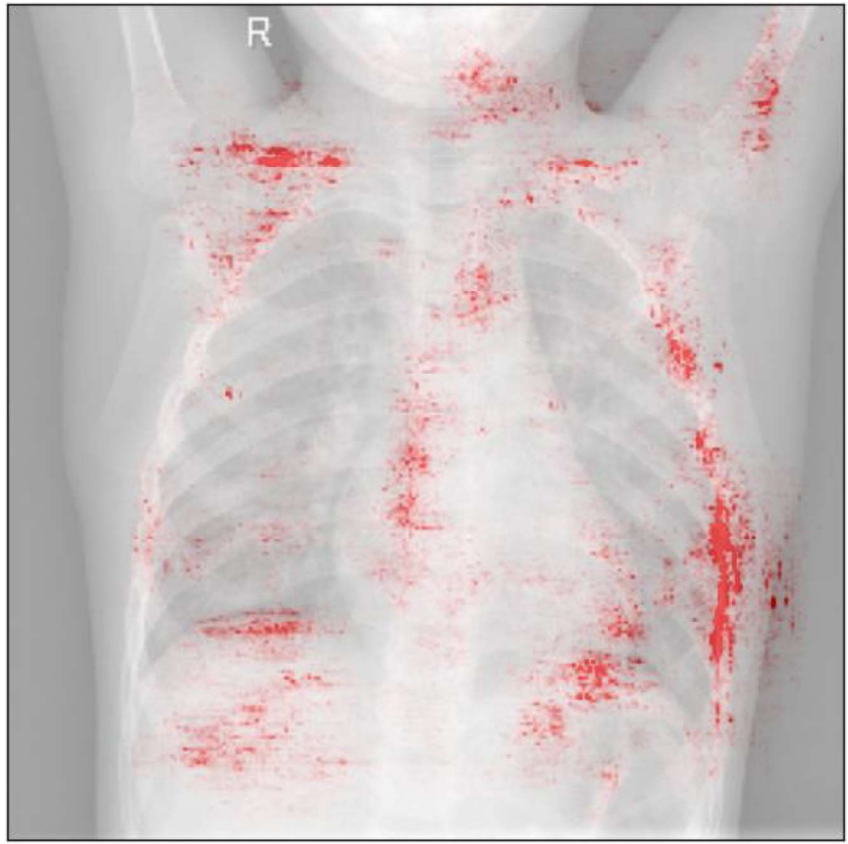

*Figure 4. The attribution of each pixel of an X-Ray image from a normal pediatric patient is represented using Integrated Gradient. The pixels of the X-Ray with high attribution to the model prediction are shown in red. The model, which was trained in the presence of a batch effect, incorrectly classifies this sample as pneumonia, and this prediction has been made based on anatomical structures and body position, which is substantially different between pediatric and adult patients. This indicates that the model learns to classify samples based on their anatomical structure, not the presence of pneumonia.*

## Discussion

With the countless proof of concepts highlighting the potential of ML and DL approaches for medical image analysis, the natural expectation is the widespread use of these approaches in clinical settings. However, when applied prospectively, the lack of generalizability is the main challenge facing these technologies. In this paper, we investigated and highlighted some of the key methodological errors that lead to models that suffer from a lack of generalizability despite achieving deceptively promising results during internal evaluation. These errors, if made, may be difficult to capture by readers, reviewers, or authors. When some methodological details are not presented, it might be impossible to detect these errors. Although not generalizable, these models define erroneous state-of-the-art models that often cannot be outperformed by generalizable models. Therefore, understanding how these methodological errors occur is essential to the reader, reviewers, and authors of ML and DL approaches.

The performance of most machine learning models is evaluated using an internal evaluation, in which the available data is partitioned into training, validation, and test sets. To achieve an unbiased estimate of performance measures for a model, the data from the training and test sets must be independent. When this assumption is violated, the internal test set does not provide an unbiased estimate of the generalization error. Therefore, any violation of the independence assumption should be avoided.

Our results demonstrated that when oversampling was incorrectly applied to the HNSCC dataset, the model achieved superficially high performance measures; however, a correct approach led to poor results as expected due to the very small number of local recurrences in the dataset. When oversampling is conducted first, and then data is randomly split into training, validation, and test sets, the same copy of a data point could appear in training and test sets. Therefore, the training and test set are no longer independent.

Similar superficial boosts were observed for incorrect application of data augmentation. When data augmentation is applied to an image, some of its characteristics change. However, there will still be many characteristics that the original image and the augmented one will share. If data augmentation is applied before data splitting, just as with oversampling, these highly correlated samples can be spread across the training, validation, and test sets. Therefore, samples highly similar to those observed during model training may be seen again in the testing or validation phases, potentially leading to high performance on the internal test set but with poor performance on external data.

Often, there are several data points associated with each patient. For instance, extracting image patches is a common practice when analyzing histopathology or 3D images (28). These patches might share some characteristics irrelevant to the study goal. Distributing the different patches derived from a single patient between training, validation, and test sets could artificially boost model performance. The resulting model will not be generalizable when applied to external data. Also, whenever there exist several data points—e.g., several MRI images—for a given patient, those data points should be assigned to only training, validation, or test set. For example, one should not assign a T2W sequence of a patient to the training and the corresponding T1W sequence of the same patient to the test set.

In radiomics, there are often a large number of features representing the statistical characteristics, shape, and texture of a region of interest. Feature selection is an important step in developing machine learning models with high-dimensional features. If feature selection is applied before splitting data, the information from all samples in the dataset is used to select a subset of features that work best for all samples. This partially exposes the test set, which will be selected in the next step, to the model and breaks the independence assumption. In this work, we observed that exposing the test samples to the feature selection methods can lead to a superficial boost in performance measures. In such cases, the selected features are chosen to be discriminative for the given test set. These features are often less discriminative when applied to unseen data. Consequently, this leads to the degradation of performance measures and a lack of generalizability.

Another key consideration in developing models is the choice of performance indicators. Such a choice should be made with the utility of the predictive model in mind. For example, using accuracy for a diagnostic model for a rare condition is often misleading, as a model that disregards all cases of the rare condition in a dataset with high class imbalance still achieves high accuracy. In addition, the cost associated with misclassification should be considered in model development. Consider a diagnostic test for a life-threatening condition. A method with high accuracy but with low recall (sensitivity) cannot be used in such situations. On the other hand, false-positive results that could lead to highly invasive procedures are prohibitive for the deployment of predictive models in clinical settings. In such scenarios, precision should also be considered as a performance indicator to guide the development and evaluation of predictive models.

Dice score and IoU are commonly used as performance measures for evaluating segmentation models. From a mathematical perspective, the Dice score is always larger than or equal to IoU (see Supplementary Materials for a mathematical proof), which encourages reporting the Dice score as the metric for evaluating segmentation models. In our example, we observed that both of these metrics achieved a high value (IoU: 0.88 and Dice: 0.94), despite obvious flaws in the segmentation of the lung. Therefore, we encourage the visual inspection of the outcome of a segmentation model as a qualitative analysis. Pixel-level accuracy, as another measure, should be avoided when evaluating small regions/volumes of interest.

It is desirable to collect and analyze imaging data from different sites, which may lead to a variety of technical differences, such as the settings used for the medical devices for acquiring diagnostic imaging (29, 30). This is beneficial for increasing sample size and generalizability of models as the data used for model development better represent the condition under study. However, if the class distribution of samples from different sites substantially varies, the aggregated dataset and the resulting models could suffer from a batch effect. It should be noted that disregarding batch effect is not rare. In a study of machine learning models used for COVID-19 diagnosis, Roberts *et al.* reported that some studies had used healthy images from pediatric patients while the COVID-19 samples came from adults (31).

Besides avoiding the aforementioned methodological errors, there are other considerations and challenges in the ML and DL domain, such as data quality and availability, bias, and explainability, which are beyond the scope of this paper. Other recent literature covers the potential of AI for misuse and provides suggestions and guidelines for how AI research can be utilized responsibly (31, 32). In addition, other current literature offers guidelines for presenting ML and DL research to ensure the reproducibility of the results (14). We would also recommend consideration of this literature by any researcher who wishes to read, review, develop, or utilize ML and DL models. The recommendations in this paper are complementary to these works. Figure 5 presents a guideline to avoid the methodological errors covered in this paper.

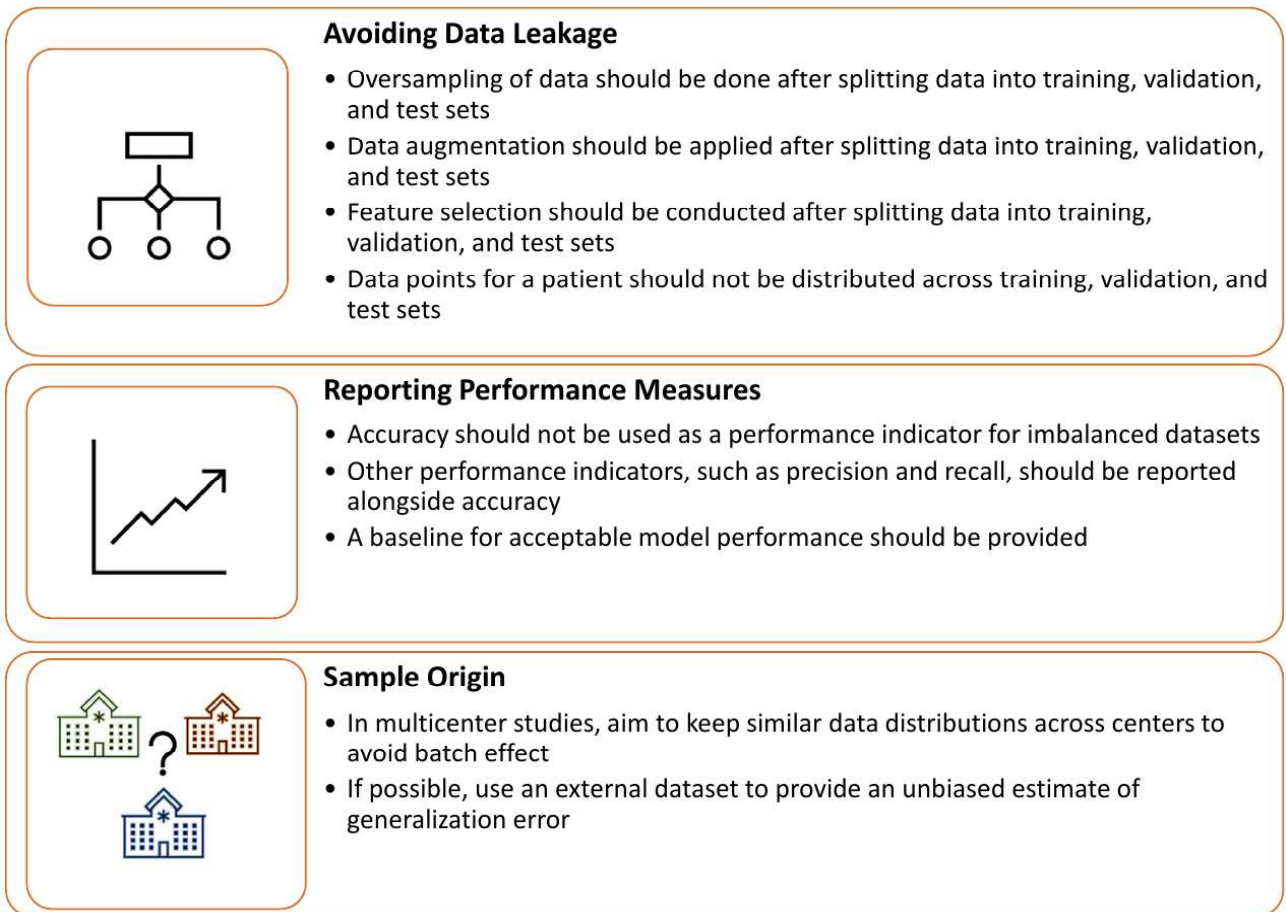

*Figure 5. Methodological guidelines for developing generalizable ML and DL models.*

Compared to natural images, medical images are often of high dimension due to their high resolutions—e.g., in histopathology whole slide images—or their 3D nature—e.g., in MRI and CT images. Consequently, analyzing such data is challenging, specifically for domains where large-scale annotated datasets are unavailable. Among the challenges posed by high data dimensionality in medical imaging are the impracticality of whole image analysis, the demanding nature of data annotation for medical images, and the increased possibility of overfitting. Due to a limited amount of VRAM (video random-access memory) offered by GPUs (Graphics Processing Units), often, whole image analysis of these high-dimensional data is impractical. Therefore, patch-based approaches are used to analyze 3D and microscopic images, which in turn poses the challenge of combining the predictions made for these patches to achieve an image-level or patient-level outcome or prediction. This also complicates the data analysis pipeline and exposes the developed models to methodological pitfalls that might affect model generalizability. Pixel/voxel level annotation for medical images is also more tedious and time-consuming compared to natural images, as annotating a single 3D image often requires manual processing of hundreds of 2D slices. Also, microscopic images often have high resolutions, e.g., 100,000 by 100,000 pixels, contributing to more annotation effort required per image. The increased possibility of model overfitting for high-dimensional data in the absence of large-scale datasets is also a major challenge affecting the generalizability of models developed for analyzing such data. Due to these challenges, following best practices and avoiding methodological pitfalls are essential for developing generalizable models with the potential to be deployed in clinical settings. Further, direct translations of methodologies developed for analyzing natural images might be impractical or suffer from a lack of generalizability. Therefore, developing methods tailored to medical images remains an active research area.

Medical image analysis is interdisciplinary, which requires contribution from both imaging, computational and medical experts. Lack of expertise in one of these domains might lead to developing models that suffer from a lack of generalizability. For example, if medical expertise is present, it is unlikely that a comparison between pediatric samples and adult patients would be considered a valid experimental design as there are substantial differences in the anatomical and imaging components between these two groups of patients. Furthermore, a model built to classify COVID-19 versus normal using lung X-rays would immediately be recognized by a medical expert as requiring a more rigorous evaluation to ensure the model does not falsely detect other lung abnormalities as COVID-19. Collaboration and cooperation amongst various experts at each stage of medical image analysis is essential for the development of models that can ultimately be applied in a clinical setting.

There are a large number of deep learning architectures available that one could customize for medical image analysis. The models used in this study are limited to the well-established architectures widely used for medical image analysis. This study could be further augmented by considering other model architectures and how susceptible they are to the violation of the independence assumption. While we expect to see variations in the performance of different DL models, considering the high capacity (26) of DL models and the commonly small sample sizes used for medical image analysis, the violation of the independence assumption is still likely to lead to results that are not generalizable. Therefore, these methodological pitfalls should be avoided regardless of the model architecture used.

Another important factor affecting the generalizability of ML and DL models, but not covered in this study, is the sample size used for model training and evaluation. Although there is general consensus on the positive impact of larger sample sizes and more varied datasets for training ML models, the literature could benefit from additional empirical analyses of the effect of sample size on model generalizability for future research.

Another factor that has been reported to affect the performance of deep learning models (33, 34), but not covered in this study, is image resolution. Using the NIH ChestX-ray14 dataset, Sabottke and Spieler (33) studied the effect of image resolution on the performance of two widely used deep learning model architectures: ResNet34 (35) and DenseNet121 (33, 36). Investigating different image resolutions ranging from $32 \times 32$ to $600 \times 600$ pixels, they showed that the optimal selection of image resolution is task-dependent and essential for increasing model performance in several classification tasks.

Lastly, in this investigation, we mainly focussed on the methodological errors resulting from the violation of the independence assumption. However, another critical requirement for developing generalizable models is that the datasets used for model training and evaluation should represent the distribution of the data in a real-world setting, i.e., data at the deployment phase. In some cases, the distribution of data used for training and evaluation of a model might not be the same. For example, samples from minority classes might be absent or poorly represented in the test set. In such a case, a model that works well on prevalent classes but works poorly on minority classes still achieves high performance measures when evaluated using such a test set. While techniques such as stratified data splitting could be effectively used to ensure similar representations for known classes or covariates across training, validation, and test sets, this issue could not be addressed through these methods for unidentified covariates. The problem is more pronounced when using small sample sizes. Therefore, generalizable models that could be reliably deployed in a clinical setting need to be evaluated using large sample sizes. This highlights the need for developing large and diverse datasets for medical image analyses.

Even for cases where training, validation, and test sets have the same distribution, the performance of a model should be regularly and systematically monitored. Due to the dynamic nature of data in real-world settings, the data distribution might change as time passes. This phenomenon—referred to as distribution shift, domain shift, or domain drift—might happen due to factors such as changes in imaging hardware, software, or protocols. In such cases, the performance of a model might degrade as time passes. This highlights that training and evaluating ML and DL models is a non-stationary and iterative process.

Due to patient privacy and intellectual property limitations, unrestricted access to data and code is often unattainable. In the absence of unrestricted access to data and code, following best practices in developing machine learning models and adhering to explicit guidelines for reporting the methodology and results are essential for developing reproducible and generalizable machine learning models in the medical domain.

In conclusion, in this research, we empirically studied several methodological pitfalls in developing ML and DL models. These pitfalls cannot be detected in an internal evaluation of models, leading to an over-optimistic estimation of model performance and consequently a lack of generalizability. Awareness of these pitfalls is important for developing generalizable ML and DL models. The suggested guidelines for avoiding these pitfalls should be considered when developing generalizable machine learning models in the medical domain.

# Appendix A

**Conventional Radiomics Analysis**: For radiomics analysis, 1652 features were extracted for each tumor in the HNSCC dataset using the *pyradiomics* package (37), which is an open-source python package for extracting radiomics features from medical images. The extracted features include shape-based, first-order statistics, gray level co-occurrence matrix, gray level run length matrix, gray level size zone matrix, gray level difference matrix, neighbourhood grey tone difference matrix, and gray level dependence matrix.

We split the data into train, validation, and test sets in a stratified manner to preserve class distribution across these sets. The data splitting was accomplished based on 3-fold outer and 3-fold inner nested cross-validation. To deal with class imbalance, we used oversampling on samples in the training set. We used a sequence of feature selection operations to deal with the high dimensionality of the radiomic features. First, we removed all constant features, as they do not offer any predictive value for model building. Then we used a univariate feature selection approach to select the top 100 features. Next, we followed a feature elimination approach to select the top 10 features for model building. Using the resulting 10 radiomic features, we built a random forest model for endpoint prediction. To achieve statistically reliable performance measures, we ran this pipeline 100 times and reported the average performance measures.

We used the *scikit-learn* Python package version 1.0.1 for feature selection and building random forest models. From the *feature_selection* module, we used *VarianceThreshold* for removing constant features, i.e., features with zero variance. Also, we used *SelectKBest* for selecting the 100 best features, where we used the *f_classif* function to assign a score to each feature. The *RFECV* with a support vector classifier (*SVC*) were used for recursive feature elimination to select the final 10 features used for model building. We used *RandomForestClassifier* from the *ensemble* module for building the random forest models. The hyperparameters were tuned using *GridSearchCV*, where *n_estimators* = 1000, *bootstrap* = True, "min_num_features" = 10, and *class_weight* = "balanced" were fixed and *max_depth* = [6, 7] and *min_samples_leaf* = [3, 4, 5] was searched.

# Appendix B

**Deep Learning Image Analysis**: For deep learning analyses, we utilized a ResNet-50 architecture[24] pretrained on ImageNet (38). The classifier layer of the ResNet model was replaced to build a binary classifier. In addition, a Drop Out layer (39) with a probability of 0.5 was added after the ResNet-50 backbone and before the classifier layer. A cross-entropy loss was used for all experiments (40). We also used Adam optimizer (41) with a learning rate of 0.0001, $\beta_1 = 0.9$, and $\beta_2 = 0.999$. A batch size of 32 was used for all experiments. All models were trained for 100 epochs, and the model with the lowest loss was selected as the best model. The Albumentations package version 0.4.5 was used for image augmentation (42). We conducted all experiments using Python 3.7 and PyTorch version 1.6 on a Titan RTX GPU machine. Figure 6 illustrates a schematic view of a DL pipeline for image classification.

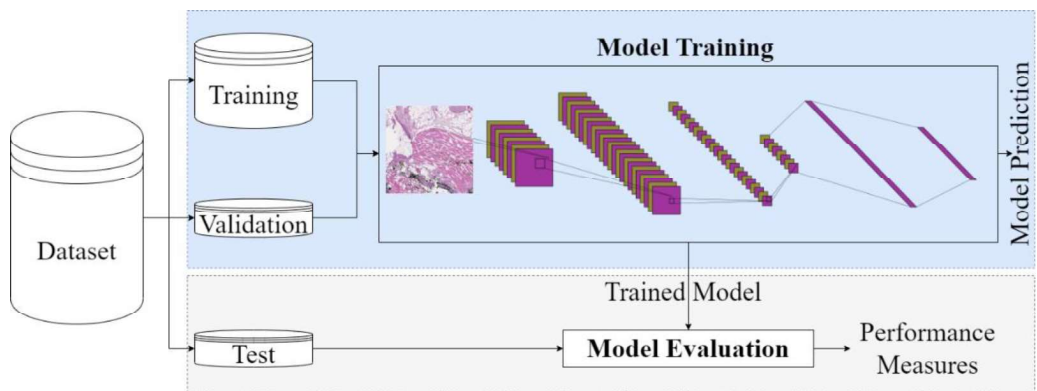

*Figure 6. A schematic view of a deep learning pipeline for image analysis.*

# Supplementary Material

## A. Dice score is always larger than or equal to IOU: a mathematical proof

Here we show that Dice score is always greater than or equal to IoU. For arbitrary sets A and B, we have:
$$\|A\| - \|A \cap B\| \geq 0 \text{ and } \|B\| - \|A \cap B\| \geq 0$$
$$(\|A\| - \|A \cap B\|) + (\|B\| - \|A \cap B\|) \geq 0$$

Where, $\|X\|$ represents the size of X. Multiplying both sides of the inequality by a positive number $\|A \cap B\|$ gives us the following:
$$\|A \cap B\| \times \|A\| - \|A \cap B\|^2 + \|A \cap B\| \times \|B\| - \|A \cap B\|^2 \geq 0$$
$$\|A \cap B\| \times \|A\| + \|A \cap B\| \times \|B\| - 2\|A \cap B\|^2 \geq 0$$

We add $\|A \cap B\| \times \|A\| + \|A \cap B\| \times \|B\|$ to both sides of the inequality.
$$2\|A \cap B\| \times \|A\| + 2\|A \cap B\| \times \|B\| - 2\|A \cap B\|^2 \geq \|A \cap B\| \times \|A\| + \|A \cap B\| \times \|B\|$$
$$2\|A \cap B\|(\|A\| + \|B\| - \|A \cap B\|) \geq \|A \cap B\| \times \|A\| + \|A \cap B\| \times \|B\|$$

Since $\|A\| + \|B\| - \|A \cap B\| = \|A \cup B\|$, we have the following:
$$2\|A \cap B\| \times \|A \cup B\| \geq \|A \cap B\| \times \|A\| + \|A \cap B\| \times \|B\|$$
$$2\|A \cap B\| \times \|A \cup B\| \geq \|A \cap B\| \times (\|A\| + \|B\|)$$
$$\frac{2\|A \cap B\|}{\|A\| + \|B\|} \geq \frac{\|A \cap B\|}{\|A \cup B\|}$$

Therefore, **Dice**$(A,B) \geq $ **IoU**$(A,B)$

## B. Checklist for avoiding methodological pitfalls covered in this paper

*Table 4. Guidelines for avoiding methodological pitfalls covered in this paper.*

| | |
|---|---|
| Oversampling of data is done after splitting data into training, validation, and test sets. | ☐ |
| Data augmentation is applied after splitting data into training, validation, and test sets. | ☐ |
| Feature selection is conducted after splitting data into training, validation, and test sets. | ☐ |
| Data points for a patient are not distributed across training, validation, and test sets. | ☐ |
| Accuracy is not used as a performance indicator for imbalanced datasets. | ☐ |
| Other performance indicators, such as precision and recall, are reported alongside accuracy. | ☐ |
| A baseline for acceptable model performance is provided. | ☐ |
| An external dataset is used to provide an unbiased estimate of generalization error (if possible). | ☐ |
| In multicenter studies, similar data distributions across centers are kept to avoid batch effect. | ☐ |

## C. The pipeline for radiomic analysis

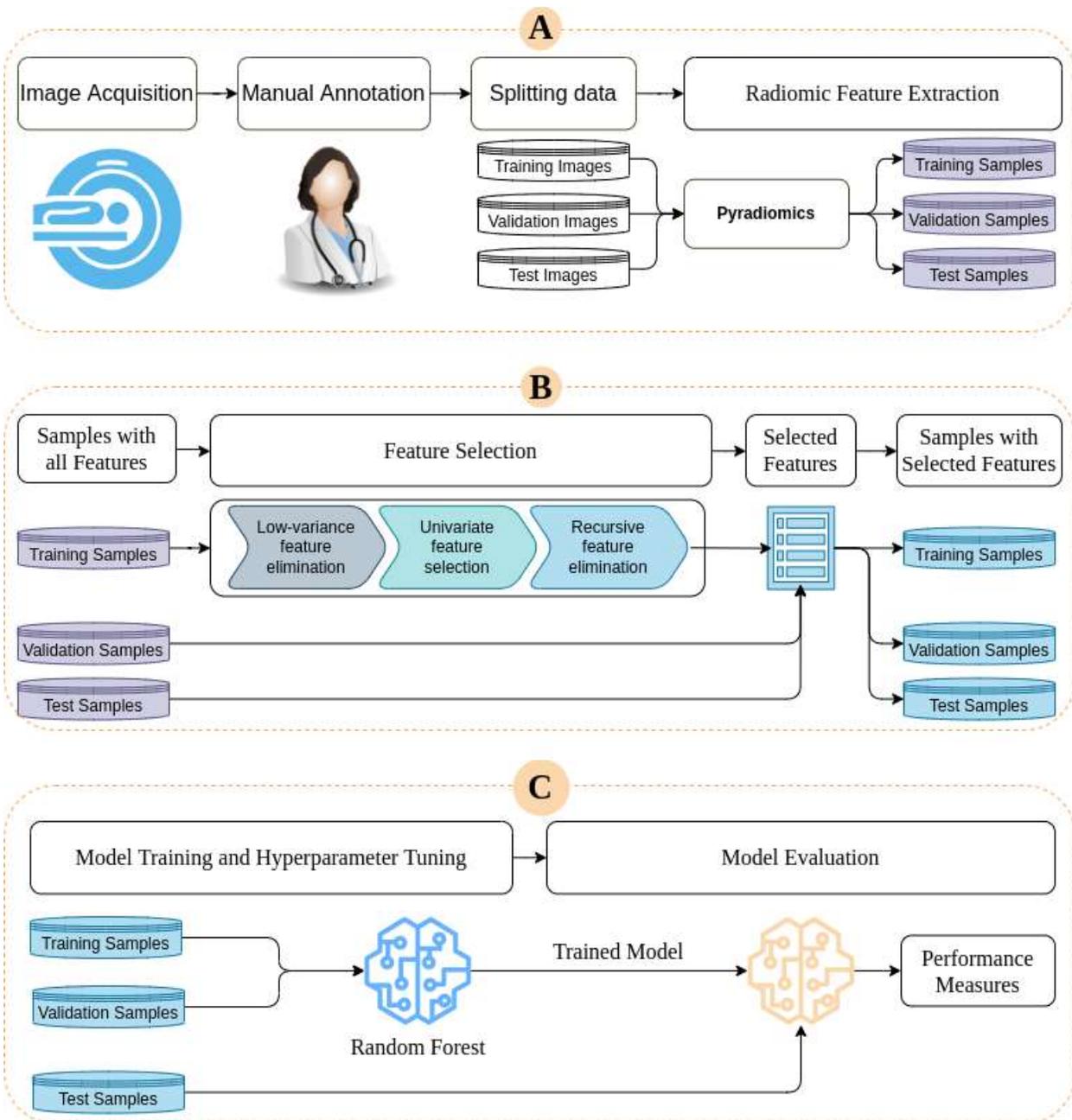

*Figure 7. The pipeline for radiomic analysis. Part A illustrates the steps from image acquisition to extracting radiomic features. The acquired images are manually contoured to annotate tumor volumes. Then images are split into training, validation, and test sets such that data for each patient is assigned to only one of these sets. Next, the Pyradiomic Python package is used to extract radiomic features for each image. Part B depicts the process for feature extraction, where training samples are used to select the most discriminative features for the task at hand. The list of selected features is then used to reduce the number of features in validation and test sets to those in the list. Part C illustrates model training and evaluation. The training set is used for model development and hyperparameter tuning, and the test set is used for model evaluation.*